\documentclass[10pt,twocolumn,letterpaper]{article}

\usepackage{wacv}
\usepackage{times}
\usepackage{epsfig}
\usepackage{graphicx}
\usepackage{amsmath}
\usepackage{amssymb}

\usepackage{amsmath}
\usepackage{amssymb}
\usepackage{array}
\usepackage{booktabs}
\usepackage{enumitem}
\usepackage{fancyvrb} 
\usepackage{graphicx}
\usepackage{makecell}
\usepackage{multirow}
\usepackage{xspace}
\usepackage{xcolor}

\DeclareMathOperator*{\argmax}{arg\,max}

\newcommand{\miro}{\textsc{TReCS}\xspace} 
\newcommand{\cocothing}{\texttt{\small thing}\xspace}
\newcommand{\cocostuff}{\texttt{\small stuff}\xspace}
\newcommand{\cocodataset}{COCO\xspace}

\newcommand{\compareexamplesmall}[7]{
\scriptsize{#1} & 
\raisebox{-.5\height}{\includegraphics[width=0.1\textwidth,height=0.1\textwidth]{figures/example_images/original/#2.jpg}} &
\raisebox{-.5\height}{\includegraphics[width=0.1\textwidth,height=0.1\textwidth]{figures/example_images/attngan/#3.png}} &
\raisebox{-.5\height}{\includegraphics[width=0.1\textwidth,height=0.1\textwidth]{figures/example_images/miro/#3.png}}
}

\def\wacvPaperID{704} 

\wacvfinalcopy 

\ifwacvfinal
\def\assignedStartPage{1} 
\fi

\begin{document}

\title{Text-to-Image Generation Grounded by Fine-Grained User Attention}

\author{Jing Yu Koh\thanks{Work done as a member of the AI Residency program.}, Jason Baldridge, Honglak Lee and Yinfei Yang \\
  Google Research \\
  \tt\small \{jykoh,jridge,honglak,yinfeiy\}@google.com\\
}

\maketitle

\begin{abstract}
Localized Narratives \cite{pont2019connecting} is a dataset with detailed natural language descriptions of images paired with mouse traces that provide a sparse, fine-grained visual grounding for phrases. We propose \miro, a sequential model that exploits this grounding to generate images. \miro uses descriptions to retrieve segmentation masks and predict object labels aligned with mouse traces. These alignments are used to select and position masks to generate a fully covered segmentation canvas; the final image is produced by a segmentation-to-image generator using this canvas. This multi-step, retrieval-based approach outperforms existing direct text-to-image generation models on both automatic metrics and human evaluations: overall, its generated images are more photo-realistic and better match descriptions.
\end{abstract}

\section{Introduction}
Text-to-image synthesis goes back at least to 1983 \cite{adorni-di-manzo-1983-natural}. WordsEye \cite{coyne2001wordseye} marked a significant evolution: it was a retrieval-based system that extracted relevant 3D models from a database to depict scenes described in text. More recently, deep neural networks based on Generative Adversarial Networks (GANs) \cite{goodfellow2014generative} have enabled end-to-end trainable photo-realistic text-to-image generation \cite{reed2016generative,zhang2017stackgan,xu2018attngan}. In contrast with these end-to-end models, others proposed hierarchical models \cite{hong2018inferring,tan2018text2scene,li2019object}; in these, object bounding boxes and segmentation masks are used as intermediate representations for realistic image generation. Several other approaches also use intermediate scene graph representations \cite{johnson2018image,yikang2019pastegan} for improving image synthesis. Others have explored dialogue-based interactions in which users incrementally provide instructions to refine and adjust generated scenes \cite{sharma2018chatpainter,el2019tell}. This provides users with greater control by allowing them to designate the relative positions of objects in the scene. However, the language used is restricted, and the images are synthetic 3D visualizations or cartoons.


\begin{figure}[!t]
    \centering
    \includegraphics[width=\linewidth]{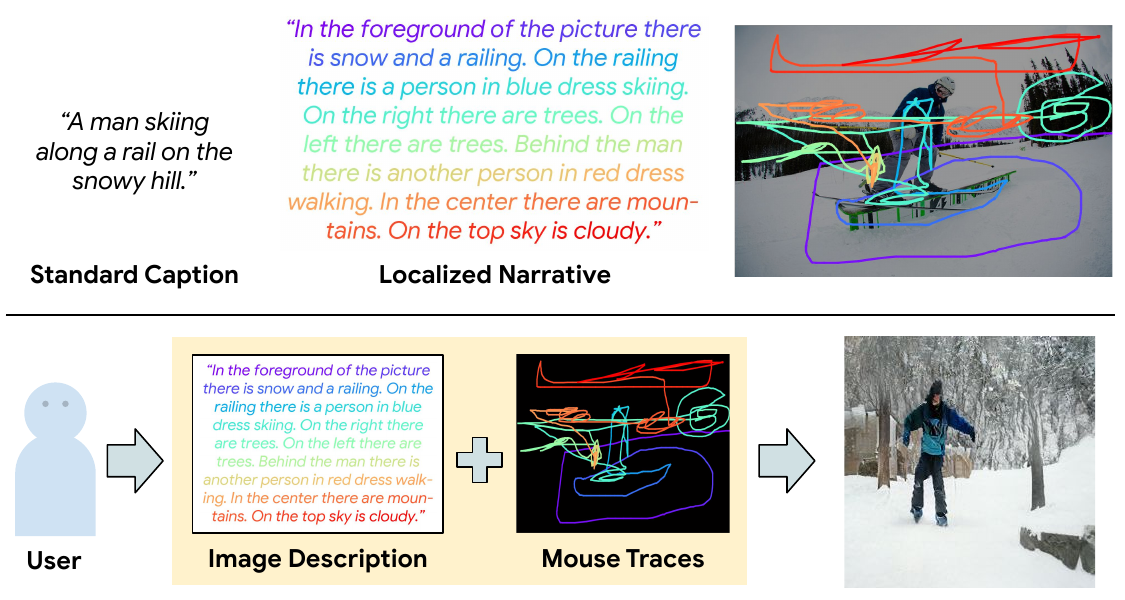}
    \caption{\textbf{[Top]}: Localized narratives comparing against standard captions. \textbf{[Bottom]}: Image synthesis from descriptions and traces.}
    \label{fig:small_model}
\end{figure}


The Localized Narratives dataset \cite{pont2019connecting} provides an alternative paradigm. Instead of writing short captions, annotators scan a mouse pointer over images while describing them. They transcribe their speech, allowing the text and traces to be time-aligned (Figure~\ref{fig:small_model}, top). These grounded narratives support the task of \textit{user attention grounded text-to-image generation} \cite{pont2019connecting}: generate an image given a free-form narrative and aligned traces (Figure~\ref{fig:small_model}, bottom).

\begin{figure*}[t]
    \centering
    \includegraphics[width=1.0\textwidth]{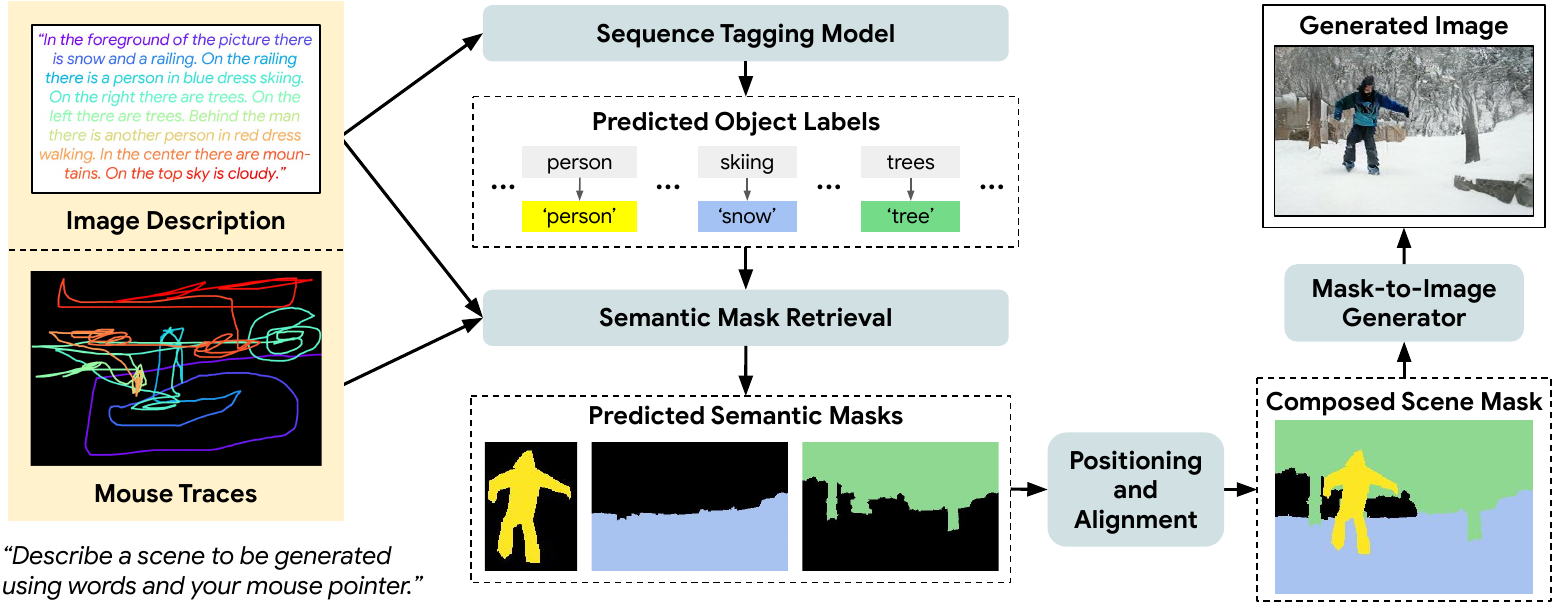}
    \caption{Multi-stage \miro\ system for image synthesis using both descriptions and mouse traces.}
    \label{fig:model}
\end{figure*}

Pont-Tuset et al.~(2020)~\cite{pont2019connecting} present a proof of concept method which naively generates images using exact matches between words and labels; we build on this core approach. Our system, \miro (\textbf{T}ag-\textbf{Re}trieve-\textbf{C}ompose-\textbf{S}ynthesize, Fig.~\ref{fig:model}), significantly enhances the image generation process by improving how language evokes image elements and how traces inform their placement.

\begin{itemize} 

\item A \textbf{tagger} predicts object labels for \textit{every} word. We train a BERT \cite{devlin2019bert} model on the output of a constrained Hidden Markov Model (HMM). The HMM is generated from noisy narrative-to-image alignments. 

\item A text-to-image dual encoder \textbf{retrieves} images with \textit{semantically relevant} masks. This helps select contextually appropriate masks for each object (e.g. skiers instead of bikers for the \textit{person} class in Fig.~\ref{fig:model}). One mask is chosen per trace sequence to maximize spatial overlap. This overcomes the lack of ground-truth text-to-object information and better grounds descriptions. 

\item Selected masks are \textbf{composed} corresponding to trace order, with separate canvases for background and foreground objects (closely related to \cocostuff and \cocothing masks from \cite{caesar2018coco}). The foreground mask is placed over the background one to create a full scene segmentation.

\item Finally, a realistic image is \textbf{synthesized} by inputting the complete segmentation to mask-to-image translation models such as SPADE \cite{park2019semantic} or CC-FPSE \cite{liu2019learning}.

\end{itemize}

\noindent
\miro exploits both text and mouse traces. Compared to other strategies, especially those requiring scene graphs, pointing with a mouse while talking is a more natural way for users to indicate their intent during image synthesis.

Localized Narratives are longer and more detailed than standard captions; they average 41.8 words in length, four times that of MS-COCO captions (averaging 10.5 words). This presents a major challenge for existing direct text-to-image models like AttnGAN \cite{xu2018attngan}. Our multi-stage strategy better copes with the detail and captures the spatial configurations of the described objects and backgrounds. On the \cocodataset portion of Localized Narratives (LN-COCO), \miro beats AttnGAN on automatic metrics of image quality as well as side-by-side human evaluations of both realism and image-text alignment: \miro is preferred 77\% of the time for realism (compared to AttnGAN's 23\%), and 52.4\% of the time for alignment (compared to AttnGAN's 32.0\%, with the remainder being ties or failures for both models). On the Open Images portion of Localized Narratives (LN-OpenImages), AttnGAN obtains better scores on automatic metrics, but \miro's output is still judged better in human evaluations, with 77.2\% of \miro images preferred for realism (compared to AttnGAN's 22.8\%), and 45.8\% preferred for language alignment (compared to AttnGAN's 40.5\%). We show that high quality images can be generated for many narratives, but there is nevertheless much room for improvement on this challenging task.


Our key contributions are:
\begin{itemize}
    \item We show, for the first time (to the best of our knowledge), viability for the very difficult task of grounded text-to-image synthesis for \textit{narratives} (as compared to prior work on shorter captions).
    \item We propose \miro, a sequential generation model that uses state-of-the-art language and vision techniques to generate high quality images that are aligned with both language and spatial mouse traces.
    \item We conduct both automatic and human evaluations that demonstrate the improved quality of \miro generated images over prior state-of-the-art. Through extensive ablative studies, we identify key components of the \miro pipeline that are essential for the user attention grounded text-to-image generation task.
\end{itemize}

\section{The \miro System}



We observed that outputs from leading end-to-end text-to-image models \cite{zhang2017stackgan,xu2018attngan,li2019object} leave much to be desired; in particular, their generated images captured a visual gist of the descriptions but lacked well-defined objects and coherent composition. Motivated by the fact that models like SPADE \cite{park2019semantic} can produce realistic images when given gold-standard segmentation masks, \cite{pont2019connecting} briefly sketch out an alternative based on mask retrieval and composition, followed by segmentation-to-image generation. We found this approach produces recognizable objects in some cases, but also often produces blank images--due largely to the limited use of the language in the narratives. Our \miro system (Figure~\ref{fig:model}) significantly enhances this strategy by better modeling the relation between language and the selection and placement of visual elements.



\subsection{Sequence Labeling with Pixel Semantics} \label{sec:pixel_semantics}

Traces in Localized Narratives cover a small portion of an image's pixels, but their fine-grained alignment to the \textit{narrative} makes them valuable indicators of the placement and scale of described items. Given both sources of information, a skilled artist could render a scene that visually captures a narrative. Datasets often used for text-to-image generation, e.g.\ \cocodataset \cite{lin2014microsoft}, Caltech-UCSD Birds \cite{welinder2010caltech}, and Oxford Flowers-102 \cite{nilsback2008automated}, do not contain such fine-grained descriptions. The latter two furthermore lack diversity in both descriptions and images.

\begin{figure}
    \centering
    \includegraphics[width=1.0\linewidth]{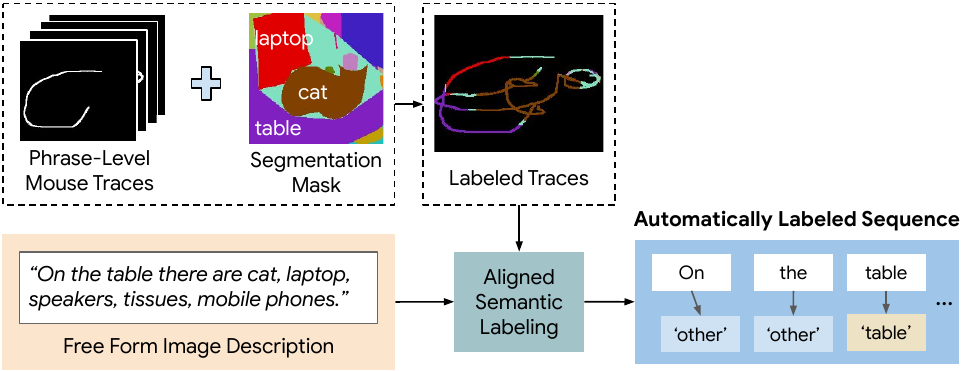}
    \caption{Overview of mouse trace sequence tagging.}
    \label{fig:seq_tagger}
\end{figure}

\miro\ exploits word-trace alignments and transition information to assign image labels to each word in training set narratives (Fig.~\ref{fig:seq_tagger}). These are used to train a BERT~\cite{devlin2019bert} model to predict tags for new narratives.
To address noise in the traces and alignments, we combine three long-standing methods to automatically refine word-object assignments: \textit{tf-idf} weighting \cite{manning:irbook}, IBM Model 1 \cite{Brown:1993}, and Hidden Markov Models \cite{Rabiner89atutorial}. We use two key observations: (1) narratives mention items that are found in the images and (2) traces pass through coherent image regions and thus provide useful category transition information (e.g., \textit{cloud} labels frequently occur next to \textit{sky} labels). To extract semantic labels of the image, we rely on the COCO-Stuff dataset \cite{caesar2018coco}, which provides pixel-level semantic segmentation masks for the \cocodataset portion of Localized Narratives.

As a starting point, we directly use word-trace alignments and gold-standard segmentation masks of each image in the training set to tag its narrative with image labels. For a given phrase and its corresponding trace, we obtain the convex hull for the trace and determine the image label that is assigned most frequently within it. This label is assigned to all words in the phrase. Using the hull rather than just the trace reduces noise, especially when annotators refer to an item by circling it with the mouse pointer (a common approach when describing objects). This produces assignments such as:

\begin{scriptsize}
\begin{Verbatim}[commandchars=\\\{\}]
this/\textcolor{teal}{snow} picture/\textcolor{gray}{other}  shows/\textcolor{teal}{snow}  a/\textcolor{teal}{snow} 
person/\textcolor{purple}{person} skiing/\textcolor{teal}{snow}  on/\textcolor{teal}{snow}  the/\textcolor{teal}{snow}  snow/\textcolor{teal}{snow} 
he/\textcolor{teal}{snow}  wore/\textcolor{purple}{person} a/\textcolor{purple}{person} helmet/\textcolor{purple}{person} on/\textcolor{purple}{person}
his/\textcolor{purple}{person} head/\textcolor{purple}{person} and/\textcolor{gray}{other} he/\textcolor{teal}{snow}  wore/\textcolor{gray}{other}
ski/\textcolor{gray}{other} pads/\textcolor{gray}{other} to/\textcolor{black}{background}
\end{Verbatim}
\end{scriptsize}

\noindent
Words do not always match their assigned labels, but the overall phrase extents correspond with traces that visited image regions with those labels. However, strong word-label associations are missed, producing evidently inappropriate tags such as \texttt{he/\textcolor{teal}{snow}} and \texttt{ski/\textcolor{gray}{other}}.





\begin{figure}
    \centering
    \includegraphics[width=1.0\linewidth]{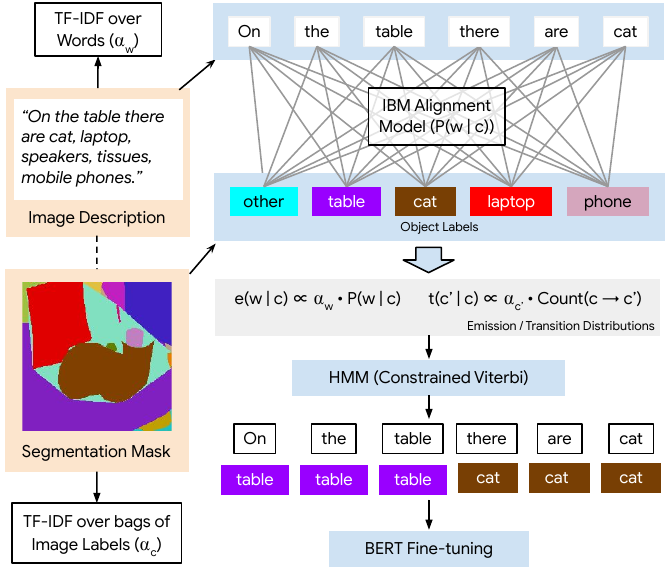}
    \caption{Semantic label refinement process.}
    \label{fig:semantic_label_refinement}
\end{figure}

\paragraph{Constrained Label Refinement} We combine these assignments with several further steps to obtain cleaner assignments (Fig.~\ref{fig:semantic_label_refinement}).  First, we compute term frequency-inverse document frequency (\textit{tf-idf}) scores $\alpha_w$ for each word $w$ in the vocabulary, treating each narrative as a document. We use these to reduce the influence of common, uninformative phrases (e.g.\ \textit{``in the image''}). \textit{tf-idf} scores $\alpha_{c}$ for image labels $c$ are computed similarly. Next, we learn word-label alignments by pairing each narrative and the corresponding bag of image labels and training IBM Model 1 on this corpus. From this, we obtain translation probabilities $P(w|c)$. 

We construct an HMM with these building blocks. The emission distributions are obtained by scaling the translation probabilities with the $\alpha_w$ scores:
\[ e(w|c) \propto \alpha_w \cdot P(w|c). \]
Similarly, the transition probabilities are defined as:
\[ t(c'|c) \propto \alpha_{c'} \cdot \text{Count}(c \rightarrow c') \]
where the counts are obtained from the noisy word-trace assignments discussed above. Add-1 smoothing is used for both distributions. Finally, we scale the contribution of the transitions: for a given step we use $e(w|c')t(c'|c)^{10}$, which we found necessary to allow tags to flip due to peaky emissions. To assign tags to a \textit{training} sentence (used to auto-supervise BERT as described in the next section), we constrain Viterbi decoding to only those image labels annotated on the image itself. The HMM produces more appropriate and semantically aligned tags; for example, the above narrative is tagged as:

\begin{scriptsize}
\begin{Verbatim}[commandchars=\\\{\}]
this/\textcolor{purple}{person} picture/\textcolor{purple}{person} shows/\textcolor{purple}{person} a/\textcolor{purple}{person}
person/\textcolor{purple}{person} skiing/\textcolor{teal}{snow} on/\textcolor{teal}{snow} the/\textcolor{teal}{snow} snow/\textcolor{teal}{snow}
he/\textcolor{purple}{person} wore/\textcolor{purple}{person} a/\textcolor{purple}{person} helmet/\textcolor{purple}{person}
on/\textcolor{purple}{person} his/\textcolor{purple}{person} head/\textcolor{purple}{person} and/\textcolor{purple}{person} he/\textcolor{purple}{person}
wore/\textcolor{purple}{person} ski/\textcolor{blue}{skis} pads/\textcolor{purple}{person} to/\textcolor{purple}{person}
\end{Verbatim}
\end{scriptsize}

\paragraph{Auto-Supervised Training} When human-annotated data is scarce, a weak generative model with strong initialization can label examples for a more powerful model that generalizes better. Garrette and Baldridge (2012) \cite{garrette2012type} show that such \textit{auto-supervised training} is effective for low-resource part-of-speech tagging. Here, we fine-tune a pretrained BERT model \cite{devlin2019bert} on the constrained HMM labels. We visually inspected the sequence tags and found that this improved both tagging quality (recall that there are no gold-standard per-word image category annotations) and final image generation quality (see detailed results in Sec. \ref{sec:seq_labeling}).

\subsection{Semantically Aligned Mask Retrieval} \label{sec:mask_retrieval}



For a narrative we wish to generate an image for, the BERT tagger detects image classes that \miro must include in the final generated image. We prepare a full semantic scene segmentation by: (1) identifying masks which match those detected classes, (2) are relevant to the narrative, and (3) are spatially aligned to the traces. 

We train a cross-modal dual encoder to retrieve $k$ training set images that best match the narrative, and then select COCO-Stuff masks of the detected classes from those images.  For each detected class instance $c_i$, we extract the masks that match that class in the top $k$ images; to satisfy spatial alignment, we select the mask $m_i$ with the highest mean intersection over union (mIOU) with the convex hull $S_i$ corresponding to the traces for instance $i$:
\[
\small
m_i = \argmax_{M_{c_i,j} \; \forall j \in \{1\ldots,k\}} \text{mIOU}(M_{c_i,j}, S_i)
\]
\noindent
where $M_{c_i,j}$ denotes mask instances of class $c_i$ for the $j^\text{th}$ retrieved image. 

\begin{figure}[t]
    \centering
    \includegraphics[width=0.9\linewidth]{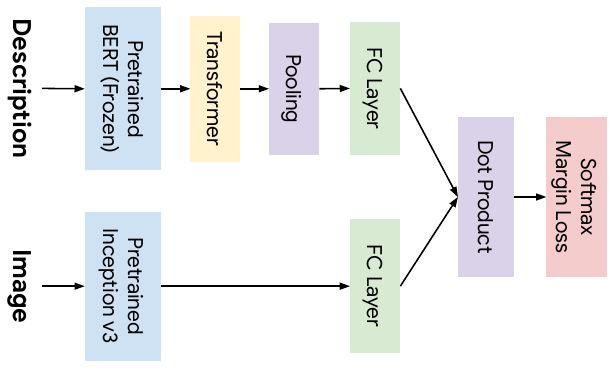}
    \caption{Dual-encoder approach for learning aligned image-text representations.}
    \label{fig:dual_encoder}
\end{figure}



We train an image-text dual encoder (Figure~\ref{fig:dual_encoder}) \cite{gillick2018end,chidambaram2019learning,yang2019improving} for cross-modal retrieval \cite{faghri2017vse++}, using the model and code of Parekh et al. (2020) \cite{crisscross_2020}. For text, we extract pre-trained BERT embeddings, which are passed into a 1-layer transformer with 8 attention heads and a hidden dimension of 128 units. The outputs are passed to a fully-connected layer which maps them to $\mathbb{R}^{2048}$. For images, we use a pre-trained Inception v3 model \cite{szegedy2016rethinking} fine-tuned during training.

Following prior work \cite{ilharco2019large,crisscross_2020}, we pre-train on Conceptual Captions \cite{sharma2018conceptual} and then train on LN-COCO (the Localized Narratives portion for MS-COCO images). The model is trained by minimizing the softmax margin loss with negatives samples from the same batch~\cite{yang2019improving}, which simulates image-caption and caption-image retrieval in each batch. 

\subsection{Mask Composition}

The masks selected in the second stage must be composed to represent the scene as a complete semantic segmentation (Fig.~\ref{fig:model}, bottom right) that is consistent with language and spatial descriptions.  Object (\cocothing) classes (e.g.\ \textit{person}, \textit{cat}, \textit{airplane}), have specific sizes, shapes, and identifiable features, while background (\cocostuff) classes (e.g.\ \textit{grass}, \textit{sky}, \textit{water}) are amorphous \cite{caesar2018coco}.
\cocostuff masks are also generally larger, so they often occlude \cocothing masks if care is not taken. Here, we use a straightforward but effective strategy that separately proposes composed layers for \cocothing (foreground) and \cocostuff (background) masks and then superimposes the former over the latter.

The \cocothing layer is created by centering each mask on its corresponding set of traces. They are placed in the reverse order they appear in the narrative---a good default strategy as annotators tend to describe salient objects before less notable details. This strategy works poorly for the \cocostuff layer because the masks are larger and clash with each other---as such, many background scenes composed this way would contain only one or two \cocostuff classes. To address this, we instead use a full semantic segmentation map from a single image---the one which has the \cocostuff objects with the highest mIOU with the \cocostuff hulls identified from the narrative and the corresponding traces. Because they are derived from a real image, these background \cocostuff masks are semantically coherent with respect to both the narrative and each other. Finally, pixels not associated with an explicit semantic mask are assigned the class label of the closest \cocostuff pixel. In addition to providing more coherent backgrounds, this also reduced the number of generated images with large blank regions.


\subsection{Mask-to-Image Translation}

The first three stages of \miro construct a novel scene segmentation respecting the descriptions and traces. The last stage aims to produce a photo-realistic image given the full scene segmentation. Compared to free-form text-to-image generation, generating from a segmentation mask is much more constrained. Mask-to-image translation is an area which has seen remarkable recent progress \cite{isola2017image,wang2018high,park2019semantic,liu2019learning}. When given a full segmentation mask for an image, existing networks are able to generate high fidelity images.

For the final stage of \miro, we experiment with two state-of-the-art mask-to-image generation models: SPADE \cite{park2019semantic} and CC-FPSE \cite{liu2019learning}, and find that CC-FPSE works better for our use case (see Section \ref{sec:image_generation_discussion}). SPADE and CC-FPSE consist of a generator and a discriminator that are trained adversarially with the GAN framework \cite{goodfellow2014generative}. The CC-FPSE generator learns a mask-to-image mapping using conditional convolution kernels; the weights of each kernel are predicted based on the mask layout. This gives the model explicit control over the generation process depending on the labels at each spatial location. The CC-FPSE discriminator incorporates multi-scale feature pyramids, which promote higher fidelity details and textures. 

Scene segmentation masks created by \miro are composed from multiple masks retrieved from different images and placed according to the narrative's traces, creating a mismatch with the gold standard segmentation masks that SPADE and CC-FPSE are trained on. Nevertheless, we find that this strategy works well compared to direct text-to-image generation, as we show in Section \ref{sec:overall_results}.

\section{Evaluation Metrics}

\label{sec:auto_metrics}

\paragraph{Image Quality} We rely foremost on side-by-side human evaluations. In each case, a human evaluator is presented with output from two competing models for the same narrative and is \textit{forced} to pick which image is more photo-realistic. (Presentation order is randomized for each judgment.) Each image is rated by 5 independent annotators to allow for a majority vote (reducing variation) and provide nuanced breakdowns. See the Appendix for details.

Inception Score (IS) \cite{salimans2016improved} is a widely-used automated metric. To compute IS, the predictions of a model (usually Inception v3 \cite{szegedy2016rethinking} pretrained on ImageNet) are obtained for a \textit{set} of images, and the distribution of its predictions is measured. Higher IS is achieved both when generated images denote clear objects and when predictions are diverse.

Fréchet Inception Distance (FID) \cite{heusel2017gans} improves on IS by comparing real and generated examples. Two multivariate Gaussians are fit to the Inception outputs of the real and generated samples; the FID score is then the Fr\'echet distance between the two Gaussians. Lower FID indicates greater similarity between the real and generated distributions.


\paragraph{Image-Language Alignment} Generated images should also match the description. \cite{hong2018inferring} assessed this fit by running a generated image through an image captioning model and computing BLEU, METEOR and CIDER scores of the generated caption given the original description. They showed that this approach correlated with human judgments. Nevertheless, this is a very indirect measure as it involves two generative models \textit{and} automated text similarity measures. More importantly, as shown in \cite{pont2019connecting}, generating long \textit{narratives} from the image alone (not including mouse traces) rather than short captions is a much harder problem, which makes caption-based measures less reliable. Hence, in this paper, we focus our evaluation for image-language alignment on human evaluations.

Similar to the human evaluation for image quality, we present generated images from two models, but also including the narrative they were conditioned on. The evaluators are asked to select the image that is more closely aligned to the narrative. Because at times neither image is a good match, evaluators can also select \textit{Neither}.
 
%

\section{Experimental Results} \label{sec:experiments}

Using automatic evaluations and human judgments, we compare \miro's performance with existing text-to-image generation models.  We evaluate all models on the \cocodataset validation set of Localized Narratives (LN-COCO) and on a held out test set of Open Images data \cite{kuznetsova2018open} that is covered by Localized Narratives (LN-OpenImages). The latter provides a stronger test of model generalization. We also perform several ablations and variations to better understand the impact of choices for each of \miro's stages.




\subsection{Main Results} 
\label{sec:overall_results}

Many models have been proposed for text-to-image synthesis \cite{hong2018inferring,xu2018attngan,li2019object,tan2019text2scene}. We compare closely to AttnGAN \cite{xu2018attngan}, as we observed that the pretrained version produced better images for LN-COCO compared to others (e.g. Obj-GAN \cite{li2019object}, see Table \ref{table:auto_results_image}). For fair comparison with \miro, we fine-tuned AttnGAN on LN-COCO's training set. Note, however, that AttnGAN uses only the narratives and not the traces---giving \miro an advantage with respect to available inputs. Incorporating traces into end-to-end models like AttnGAN is non-trivial and worth exploring in future.


\begin{figure}
\begin{center}
\includegraphics[width=0.8\linewidth]{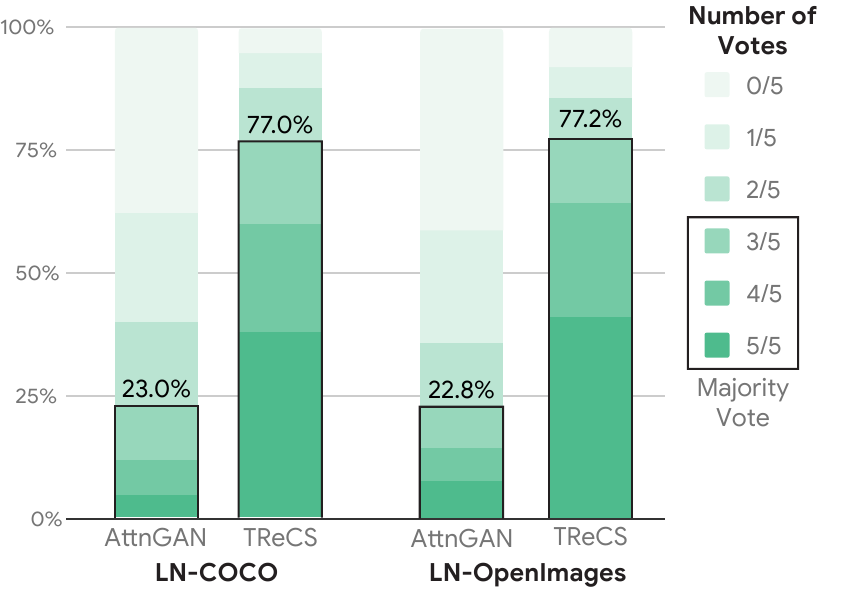}
\end{center}
\caption{Human evaluation of image quality on LN-COCO validation set and LN-OpenImages test set. Models were fine-tuned on the LN-COCO training set. Of the decisions with 5/5 votes (indicating unanimous preference), \miro was selected 88.3\% of the time on LN-COCO, compared to 11.7\% for AttnGAN. On LN-OpenImages, \miro was selected unanimously 84.1\%, compared to 15.9\% for AttnGAN.} \label{fig:realism_human_results}
\end{figure}

\begin{figure}
\begin{center}
\includegraphics[width=1.0\linewidth]{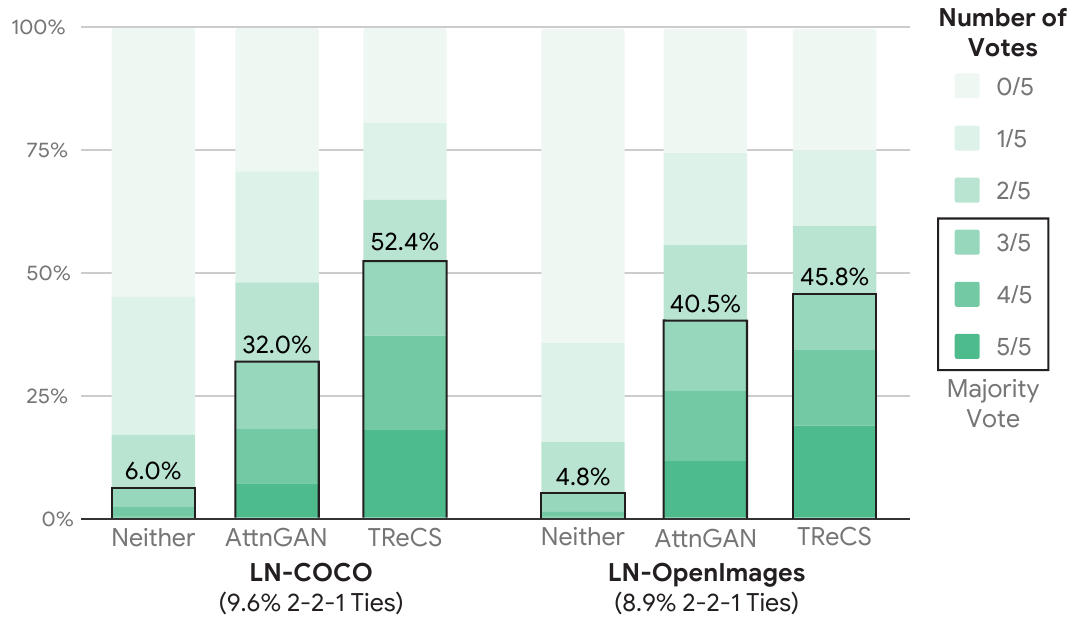}
\end{center}
\caption{Human evaluation of image-text alignment on LN-COCO validation and LN-OpenImages test sets. Models were fine-tuned on the LN-COCO training set. Of the decisions with 5/5 votes (indicating unanimous preference), \miro was selected 71.0\% of the time on LN-COCO, compared to 27.8\% for AttnGAN. On LN-OpenImages, \miro was selected unanimously 61.4\%, compared to 37.9\% for AttnGAN.} \label{fig:text_human_results}
\end{figure}

\begin{figure*}
\small
\begin{center}
\begin{tabular}{>{\arraybackslash}m{0.19\linewidth}@{\hskip 0.5em}c@{\hskip 0.1em}c@{\hskip 0.1em}c@{\hskip 0.5em}>{\arraybackslash}m{0.19\linewidth}@{\hskip 0.5em}c@{\hskip 0.1em}c@{\hskip 0.1em}c@{\hskip 0em}}
\textbf{Caption} & \textbf{Original} & \textbf{AttnGAN} & \textbf{\miro} & \textbf{Caption} & \textbf{Original} & \textbf{AttnGAN} & \textbf{\miro} \\ \midrule
\addlinespace[0em]

\compareexamplesmall{In this image a lady wearing green cap, blue jacket is skiing. She is holding two sticks. The ground is full of snow. In the background there are trees ...}{466602}{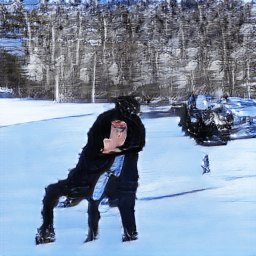}{}{}{}{} &

\compareexamplesmall{In this picture we can see a zebra on grass and eating grass and in the background we can see trees.}{552902}{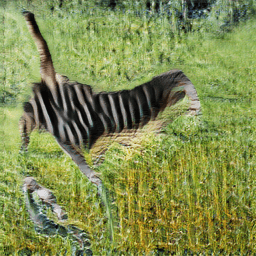}{}{}{}{} \\\addlinespace[-0.1em]

\compareexamplesmall{This is a collage of two photos. Here we can see a woman playing in the ground and she is holding a racket with her hands. And there is a mesh.}{166165}{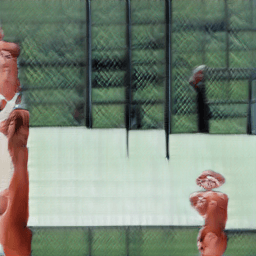}{}{}{}{} &

\compareexamplesmall{In this picture we can see a bus and a person in it. ... Some grass is seen on the ground. There are some cars and tree in the background.}{296224}{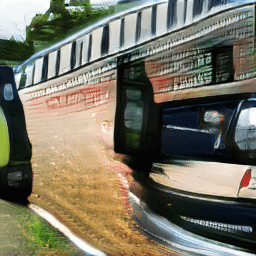}{}{}{}{} \\\addlinespace[-0.1em]

\compareexamplesmall{In this image I can see a dog is sitting in a vehicle. I can also see a tree and few plants in the background. Here I can see sky.}{65485}{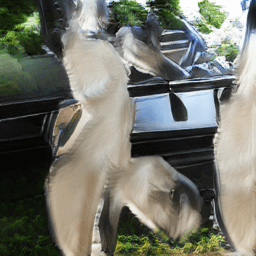}{}{}{}{} &

\compareexamplesmall{This picture describe about Street view. In the center we can see a clock tower it's like a road On the tower we can see a street name as pigerpodi is written. Behind ...}{234366}{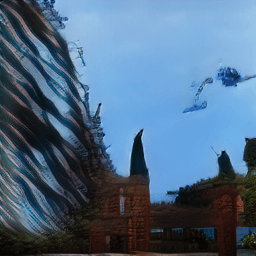}{}{}{}{} \\\addlinespace[-0.1em]

\compareexamplesmall{Here we can see a giraffe in the middle and we can see grass present in baskets on the fencing, we can see plants and trees present}{186637}{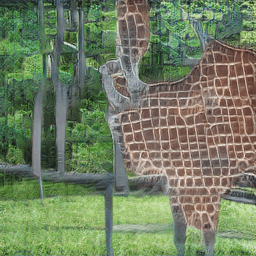}{}{}{}{} &

\compareexamplesmall{In this image we can see an elephants on the ground, here are the bamboo sticks, here is the grass, here is the rope, here are the trees, and at above here is the sky.}{83113}{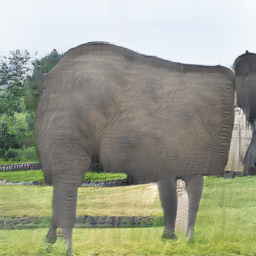}{}{}{}{} \\\addlinespace[-0.1em]

\compareexamplesmall{there is it teddy bear with spectacles with sitting on a sofa with keyboard microphone and video game.}{42889}{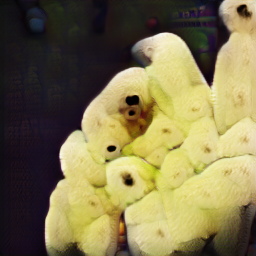}{}{}{}{} &

\compareexamplesmall{In this image there is a bench on stone slab. Bottom of image is a grassy land. Background of image there are few plants.}{322829}{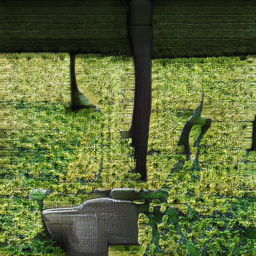}{}{}{}{}

\end{tabular}
\end{center}
\caption{Original and generated images for cherry picked examples from LN-COCO.} \label{fig:qualitative_results_realism}
\end{figure*}

\begin{figure}
\small
\begin{center}
\begin{tabular}{>{\arraybackslash}m{0.3\linewidth}@{\hskip 0.5em}c@{\hskip 0.1em}c@{\hskip 0.1em}c@{\hskip 0.5em}>{\arraybackslash}m{0.01\linewidth}@{\hskip 0.5em}c@{\hskip 0.1em}c@{\hskip 0.1em}c@{\hskip 0em}}
\textbf{Caption} & \textbf{Original} & \textbf{AttnGAN} & \textbf{\miro} \\ \midrule  \addlinespace[0em]





\compareexamplesmall{A person wearing jacket, helmet is on a ski board holding ski sticks. There is snow. On the back there is a banner, stand, ropeway ...}{421060}{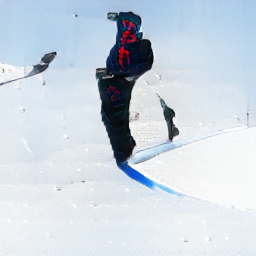}{}{}{}{} \\\addlinespace[0.1em]

\compareexamplesmall{Group of people standing and we can see kites in the air and sky with clouds. A far we can see trees. This is grass.}{163117}{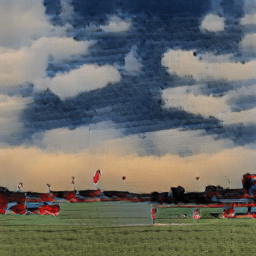}{}{}{}{} \\\addlinespace[0.1em]

\compareexamplesmall{In this picture we can see one boy is holding a bat and playing a game, he is keeping a cap. soundings there is ...}{311928}{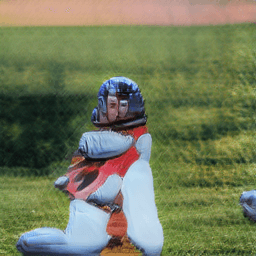}{}{}{}{} \\\addlinespace[-0.1em]

\compareexamplesmall{In the picture there is a road on the road there are many vehicles there are many poles on the road there are many trees ...}{93154}{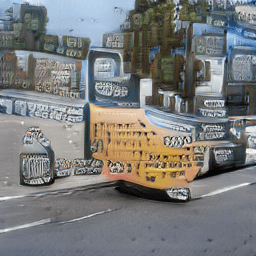}{}{}{}{} \\\addlinespace[0.1em]

\compareexamplesmall{The image is outside of the city. In the image in middle there are few bags, on right side we can see a person standing ...}{360097}{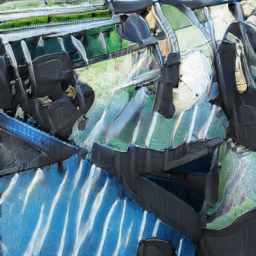}{}{}{}{} \\\addlinespace[0.1em]

\compareexamplesmall{In this image we can see three fire engines on the road. In the background there are trees, houses and sky.}{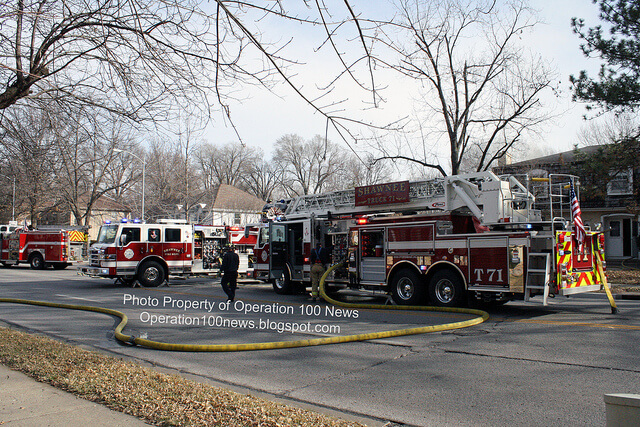}{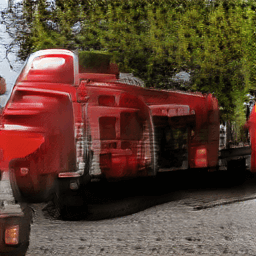}{}{}{}{} \\\addlinespace[0.1em]

\compareexamplesmall{In this image i can see few benches and the ground covered with the snow. In the background i can see few ...}{222458}{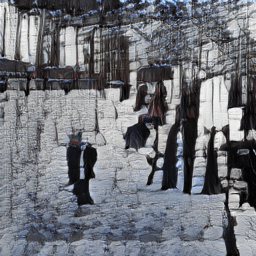}{}{}{}{} \\\addlinespace[0.1em]

\compareexamplesmall{In this image I can see the road ... In the back there are signal lights and the vehicles. I can see many trees, building ...}{553339}{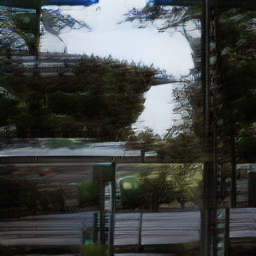}{}{}{}{} \\\addlinespace[0.1em]

\compareexamplesmall{In this image I can see a cat on a vehicle.}{530099}{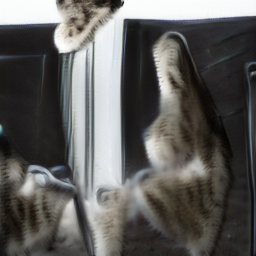}{}{}{}{} \\\addlinespace[0.1em]

\compareexamplesmall{This picture shows a man standing and skiing and we see blue cloudy sky and a tree and he wore a cap on his head}{2532}{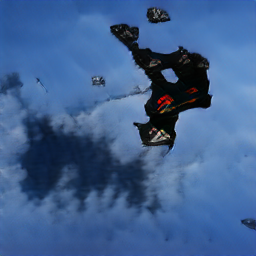}{}{}{}{} \\\addlinespace[0.1em]

\compareexamplesmall{This woman wore yellow t-shirt, headband, holding bottle and standing beside this yellow hydrant. On this grass ...}{375493}{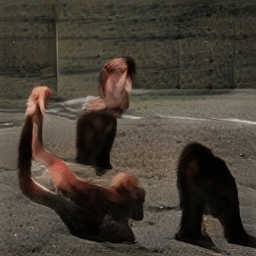}{}{}{}{} \\\addlinespace[0.1em]

\compareexamplesmall{This is the picture of a floor where we have some laptops, bags and an other back pack in which some things ...}{32610}{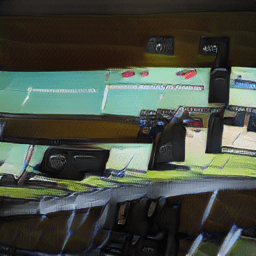}{}{}{}{} \\\addlinespace[0.1em]


\end{tabular}
\end{center}
\caption{Original and generated images for random examples from LN-COCO.} \label{fig:qualitative_results_random_small}
\end{figure}

\textbf{Image Quality}
Qualitatively, \miro's images are crisper and more realistic, as seen in cherry-picked (Fig.~\ref{fig:qualitative_results_realism}) and random (Fig.~\ref{fig:qualitative_results_random_small}) examples. AttnGAN tends to produce textures and blobs that are semantically relevant (e.g.\ giraffe patterns) but do not represent clear objects. This is confirmed in side-by-side human evaluation of image quality (Fig.~\ref{fig:realism_human_results}): for 77.0\% of 1000 LN-COCO narratives, the \miro image was preferred to the AttnGAN image. The same preference was found for the LN-OpenImages test set. We attribute the improvement in image quality to our staged approach and use of traces. By composing a scene mask instead of direct generation, \miro\ explicitly imposes object shapes when composing scenes, and subsequently take advantages of strong mask-to-image generation models to create high fidelity images.

\begin{table}
\begin{center}
\resizebox{\linewidth}{!}{%
\begin{tabular}{@{}l@{\hspace{10mm}}l@{\hspace{10mm}}r@{\hspace{4mm}}r@{}}
\textbf{Dataset} & \textbf{Method} & \textbf{IS} $\uparrow$ & \textbf{FID} $\downarrow$ \\ \midrule
\multirow{4}{*}{LN-COCO}  & Obj-GAN$^\dagger$ &  16.5 & 66.5 \\
 & AttnGAN$^\dagger$ &  17.4 & 59.4 \\
 & AttnGAN &  20.8 & 51.8 \\
 & \miro   &  \underline{21.3} & \underline{48.7} \\[1mm]
  \hline \\[-3mm]
\multirow{2}{*}{LN-OpenImages} & AttnGAN & \underline{15.3} & \underline{56.6} \\
& \miro   &  14.7 & 61.9
\end{tabular}}
\end{center}
\caption{Image quality scores on LN-COCO validation and LN-OpenImages test sets. $\uparrow (\downarrow)$ indicates that a higher (lower) number is better performance. $\dagger$ indicates models pretrained on the original \cocodataset, but not fine-tuned on the LN-COCO training set.}\label{table:auto_results_image}
\end{table}

Table \ref{table:auto_results_image} shows IS and FID scores for both models on both datasets. Clearly, fine-tuning AttnGAN on LN-COCO makes a large improvement (59.4 FID $\rightarrow$ 51.8 FID), so all other AttnGAN results in this paper are for the fine-tuned version. On LN-OpenImages, \miro and fine-tuned AttnGAN swap leadership on these measures: e.g., \miro is 3.1 FID points better on LN-COCO and 5.3 FID points worse on LN-OpenImages. We find that these metrics provide valuable feedback while developing models, but stress that human evaluation provides the better measure of generated image quality. Note that it is possible to optimize for IS directly and create incomprehensible or adversarial images that nonetheless achieve IS as high as 900 \cite{barratt2018note}.

\textbf{Image-Text Alignment}
\miro\ also outperforms AttnGAN on human evaluations of image-text alignment (see Fig.~\ref{fig:text_human_results}). Of 1000 LN-COCO held out images, 52.4\% were chosen by human raters as being better aligned to the given narrative, compared to 32.0\% of AttnGAN images. We observe a similar trend on LN-OpenImages, with \miro winning 45.8\% of contests to AttnGAN's 40.5\%.

\miro also has distinctly better performance when considering cases for which there is full agreement (i.e. decisions where 5/5 voters selected either model, or neither). For these images, \miro is selected 71.0\% of the time, as compared to 27.8\% for AttnGAN (the remaining being \textit{Neither}). Similarly, on LN-OpenImages, \miro is selected unanimously 61.4\% of the time, as compared to 37.9\% for AttnGAN, indicating a clear preference for \miro images when evaluating for text-alignment.

\miro's superior performance may be due to its ability to handle longer free form descriptions. Narratives are much longer (average of 41.8 words) than MS-COCO captions (average of 10.5 words). Also, the narratives are transcriptions of free-form speech and incorporate filler words used in everyday speech. This data presents a challenge for existing text-to-image synthesis models, which were originally created to handle concise and clean captions. By explicitly assigning image labels to words, performing cross-modal matching during mask retrieval, and composing foreground and background's separately, \miro\ captures the full range of described objects more effectively.

\subsection{Ablations} \label{sec:seq_labeling}
\begin{figure}[t]
    \centering
    \resizebox{\linewidth}{!}{%
    \small
    \begin{tabular}{@{}>{\arraybackslash}m{0.3\linewidth}>{\arraybackslash}m{0.175\linewidth}>{\arraybackslash}m{0.175\linewidth}>{\arraybackslash}m{0.175\linewidth}@{}}
    \textbf{\begin{tabular}[c]{@{}l@{}}Query\\ Description\end{tabular}} &
    \textbf{\begin{tabular}[c]{@{}c@{}}Retrieved\\ Image \#1\end{tabular}} &
    \textbf{\begin{tabular}[c]{@{}c@{}}Retrieved\\ Image \#2\end{tabular}} &
    \textbf{\begin{tabular}[c]{@{}c@{}}Retrieved\\ Image \#3\end{tabular}} \\ \midrule
    \scriptsize{In this picture we can see food and spoon in the plate.} & \includegraphics[width=1.0\linewidth,height=1.0\linewidth]{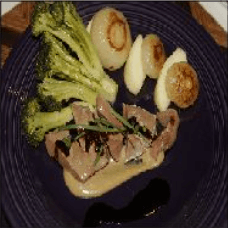} & \includegraphics[width=1.0\linewidth,height=1.0\linewidth]{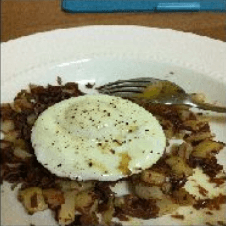} & \includegraphics[width=1.0\linewidth,height=1.0\linewidth]{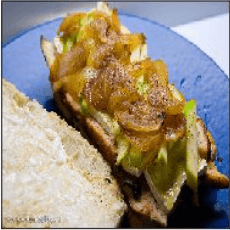} \\
    \scriptsize{In the image there is a donkey truck on the side of the road beside it there is a caution board, on back ...} & \includegraphics[width=1.0\linewidth,height=1.0\linewidth]{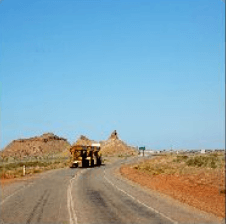} & \includegraphics[width=1.0\linewidth,height=1.0\linewidth]{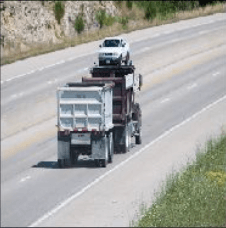} & \includegraphics[width=1.0\linewidth,height=1.0\linewidth]{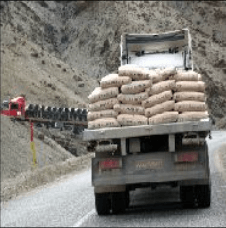} \\
    \scriptsize{In a room people are seated on wooden chairs. In the center there is a rectangular dining table ...} & \includegraphics[width=1.0\linewidth,height=1.0\linewidth]{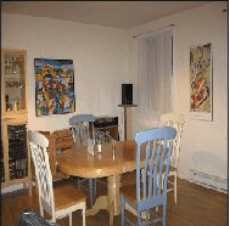} & \includegraphics[width=1.0\linewidth,height=1.0\linewidth]{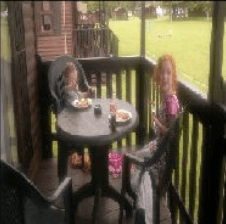} & \includegraphics[width=1.0\linewidth,height=1.0\linewidth]{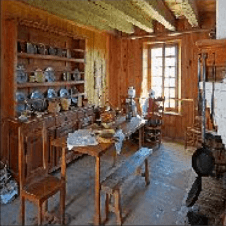}  \\
    \end{tabular}
    }
    \caption{Dual encoder retrieval examples.}
    \label{fig:retrieval_qualitative}
\end{figure}

\paragraph{Sequence Labeling} Using HMM tags (Section \ref{sec:pixel_semantics}), we fine-tuned an uncased BERT-Large model \cite{devlin2019bert}.
Weights are optimized using Adam \cite{kingma2014adam} with a learning rate of 1e-5. During training and inference, we set the class probabilities of the COCO-Stuff \texttt{other} and \texttt{background} tags to 0. This assists in downstream image generation, as we found that these classes were not meaningful when presented to the mask-to-image translation models.

Table \ref{table:ablation} shows that training BERT on the output of the HMM (auto-supervision) improves image quality over training it on noisy labels (20.0/49.9 $\rightarrow$ 21.3/48.7) or using the output of the HMM itself (20.7/49.0 $\rightarrow$ 21.3/48.7). Note that the HMM tags exploit ground-truth image-level labels (and hence are not valid for testing).
We observed from manually inspecting image outputs that auto-supervision improved both image quality and image-text alignment.

\begin{table}
\begin{center}
\resizebox{\linewidth}{!}{%
\begin{tabular}{@{}l@{\hspace{6mm}}l@{\hspace{6mm}}c@{\hspace{5mm}}c@{}}
\textbf{Sequence Labels} &
\textbf{Generator} &
\textbf{IS} $\uparrow$ &
\textbf{FID} $\downarrow$ \\ \midrule
BERT (raw labels) & CC-FPSE & 20.0 & 49.9 \\
HMM    & SPADE   & 20.2 & 50.6 \\
HMM    & CC-FPSE & 20.7 & 49.0 \\
BERT (HMM)   & SPADE   & 20.2 & 49.5 \\
BERT (HMM)  & CC-FPSE & \underline{21.3} & \underline{48.7}
\end{tabular}}
\end{center}
\caption{Ablation experiments on the validation set of LN-COCO.  $\uparrow (\downarrow)$ indicates that a higher (lower) number equates better performance. BERT (raw labels) indicate a BERT model that was trained on raw segmentation labels, as compared to the HMM processed labels. A dual-encoder with $k{=}5$ is used for mask retrieval.} \label{table:ablation}
\end{table}



\paragraph{Mask Retrieval}
The retrieval model is strong (Table \ref{table:retrieval_generative_results_combined}): given a query narrative, the groundtruth image is retrieved 53.8\% of the time in the top 5 and 95.4\% in the top 100, over 8573 images in LN-COCO's validation set. Inspecting retrieved results (e.g.\ Fig.~\ref{fig:retrieval_qualitative}) indicates that this underestimates performance: many retrieved images are good matches, but these narrative-image connections are not in the paired data. Ilharco et al. (2020) \cite{ilharco2019large} provide human evaluations that show retrieval performance is underestimated by the available data, and Parekh et al. (2020) \cite{crisscross_2020} provide new annotations that partly address this gap.

\begin{table}[t]
\begin{center}
\begin{tabular}{@{}l@{\hspace{10mm}}c@{\hspace{4mm}}c@{\hspace{4mm}}c@{\hspace{4mm}}c@{\hspace{4mm}}@{}}
\multirow{2}{*}{\textbf{$k$}} &
\multicolumn{2}{c@{\hspace{8mm}}}{\textbf{Recall@$k$(\%)}} &
\multirow{2}{*}{\textbf{IS} $\uparrow$} &
\multirow{2}{*}{\textbf{FID} $\downarrow$} \\ \cmidrule(l{-2mm}r{4mm}){2-3}
& {C$\rightarrow$I} & {I$\rightarrow$C} & & \\ \midrule
GT  & - & - & 22.0 & 44.6 \\
1   & 26.3 & 27.0 & 20.3 & 49.0 \\
5   & 53.8 & 52.2 & \underline{21.3} & \underline{48.7} \\
10  & 66.8 & 63.1 & 21.0 & 50.9 \\
20  & 78.4 & 74.1 & 20.7 & 52.3 \\
50  & 89.9 & 86.6 & 20.0 & 56.5 \\
100 & 95.4 & 92.9 & 18.8 & 60.5
\end{tabular}
\caption{Retrieval and generation evaluation (LN-COCO) with varying $k$. \textbf{Labeler:} BERT; \textbf{image sequencer:} CC-FPSE. \textbf{[C$\rightarrow$I]}: Caption to image retrieval (vice versa for \textbf{I$\rightarrow$C}). \textbf{[GT]}: Retrieval using groundtruth masks, providing IS/FID upper bounds.} \label{table:retrieval_generative_results_combined}
\end{center}
\end{table}

Increasing $k$ (number of retrieved images) improves image generation quality up to a certain point, after which IS and FID drop. For LN-COCO, $k{=}5$ produces the best images (Table \ref{table:retrieval_generative_results_combined}).
This matches our intuition: as $k$ increases, the number of less-relevant masks increases, and some of these are chosen because their shape better matches the trace hull.
Conversely, when $k$ is too small, there are insufficient semantically connected masks.

\paragraph{Image Generation} \label{sec:image_generation_discussion}
CC-FPSE \cite{liu2019learning} generally outperforms SPADE \cite{park2019semantic} on IS/FID while keeping other choices fixed (Table \ref{table:ablation}). We manually inspected generated images from both models and found CC-FPSE also had better image-text alignment. Hence, we used CC-FPSE to synthesize the primary results in Section \ref{sec:overall_results}.

We also tried fine-tuning CC-FPSE  with segmentation outputs created by our retrieval module. The model is trained on a set of original segmentation annotations and a set of noisy segmentation masks created by our retrieval module. The goal of this training process was to fine-tune the generative model to handle noise introduced in the mask retrieval process. However, this training setting did not result in any notable improvements to generation results, most likely due to the large spatial disjoint between noisy masks and real image pixels. Hence, we used the pretrained CC-FPSE model to report our final results.

\section{Conclusion}

As judged by people, \miro outperforms AttnGAN in producing higher quality images that are better aligned to the descriptions. \miro is a step towards an interface that allows a user to describe a scene while controlling size and placement naturally. Such interfaces would support various applications, allowing users to rapidly create interactive scenes without specialized software \cite{coyne2001wordseye}, assist the creative process of an artist \cite{ha2017neural}, or generate realistic images that are more closely grounded to intended descriptions \cite{sharma2018chatpainter}.

\section*{Acknowledgements}
We would like to thank Jordi Pont-Tuset, Austin Waters, Han Zhang, and many others for their insightful ideas and discussions during the development of this paper. We would also like to thank the Google Data Compute team for collecting human evaluations on our generated images. We thank the anonymous reviewers for their helpful comments.

{\small
\bibliographystyle{ieee_fullname}
\bibliography{main}
}

\end{document}